\documentclass[conference]{IEEEtran}
\IEEEoverridecommandlockouts
\usepackage{cite}
\usepackage{amsmath,amssymb,amsfonts}
\usepackage{algorithmic}
\usepackage{graphicx}
\usepackage{textcomp}
\usepackage{xcolor}

\usepackage{xcolor}
\usepackage{listings}
\usepackage{subfigure}
\usepackage{amsfonts}
\usepackage{url}
\usepackage{blindtext}
\usepackage{dirtytalk}
\usepackage{verbatim}
\usepackage{multirow}
\usepackage[super]{nth}
\usepackage{siunitx}
\usepackage{hyperref}
\usepackage{threeparttable}

\newcommand{\todo}[1]{\textcolor{red}{TODO: #1}}

\def\BibTeX{{\rm B\kern-.05em{\sc i\kern-.025em b}\kern-.08em
    T\kern-.1667em\lower.7ex\hbox{E}\kern-.125emX}}
\begin{document}

\title{Action Space Shaping in Deep Reinforcement Learning}

\author{
    \IEEEauthorblockN{Anssi Kanervisto}
    \IEEEauthorblockA{\textit{School of Computing} \\
    \textit{University of Eastern Finland}\\
    Joensuu, Finland \\
    anssk@uef.fi}
\and
    \IEEEauthorblockN{Christian Scheller}
    \IEEEauthorblockA{\textit{Institute for Data Science} \\
    \textit{University of Applied Sciences}\\
    \textit{Northwestern Switzerland}\\
    Windisch, Swizerland \\
    christian.scheller@fhnw.ch}
\and
    \IEEEauthorblockN{Ville Hautam\"aki}
    \IEEEauthorblockA{\textit{School of Computing} \\
    \textit{University of Eastern Finland}\\
    Joensuu, Finland \\
    villeh@uef.fi}
    
\thanks{
    We gratefully acknowledge the support of NVIDIA Corporation with the donation of the Titan Xp GPU used for this research.

    \textcopyright 2020 IEEE.  Personal use of this material is permitted.  Permission from IEEE must be obtained for all other uses, in any current or future media, including reprinting/republishing this material for advertising or promotional purposes, creating new collective works, for resale or redistribution to servers or lists, or reuse of any copyrighted component of this work in other works.
}}

%\IEEEpubid{\begin{minipage}{\textwidth}\ \\[12pt]
%978-1-7281-4533-4/20/\$31.00 \copyright 2020 IEEE
%\end{minipage}}

\maketitle

\begin{abstract}
    Reinforcement learning (RL) has been successful in training agents in various learning environments, including video games. 
    However, such work modifies and shrinks the action space from the game's original. 
    This is to avoid trying \say{pointless} actions and to ease the implementation.
    Currently, this is mostly done based on intuition, with little systematic research supporting the design decisions. 
    In this work, we aim to gain insight on these action space modifications by conducting extensive experiments in video game environments. 
    Our results show how domain-specific removal of actions and discretization of continuous actions can be crucial for successful learning. 
    With these insights, we hope to ease the use of RL in new environments, by clarifying what action-spaces are easy to learn.
\end{abstract}

\begin{comment}
\textcolor{red}{Things to do:
\begin{itemize}
    \item Focus: video games first, reinforcement learning second.
    \item Study previous work on the topic (for background/related work section), and generally on action spaces. \textbf{Focus on video games.}
    \item Paper on discretizing continuous action spaces: "Discretizing continuous action space for on-policy optimization"
    \item Idea for short paper after this one: Studying the frame-skip learning from imitation set.
\end{itemize}
}

\textcolor{red}{Things for the next meeting (Tue 3.3):
\begin{itemize}
    \item Anssi: Writing. Drawing plots. Why did we get these results? Check the differences between rllib and stable-baselines. \textbf{First draft ready by next meeting}. Draw the plots. Try binning of continuous actions.
    \item Christian: Finishing up experiments, writing out SC2 results.
    \item Question: We should study the actions that ruin everything, and run experiments with those environments (e.g. where single action leads to death easily. Related to that icy-cliff toy-MDP?)
\end{itemize}
}
\end{comment}


\section{Introduction}
\label{sec:intro}
    %\todo{Mention UnityML and how these tips could be used to make environments in there}
    %\todo{Stress our main point: Somebody has new env and learning algorithm tuned, but how do they know if the action space is not ruining the training? We here aim to provide some answers (multidiscrete vs discrete? Continuous actions? minimal spaces?}
    % When researchers and developers wish to use \textit{reinforcement learning} (RL)~\cite{sutton2018reinforcement} in their work, there are numerous design choices to be made before they can start training their agents.
    %Applying \textit{reinforcement learning} (RL)~\cite{sutton2018reinforcement} comes with numerous design choices to be made:
    \textit{Reinforcement learning}~\cite{sutton2018reinforcement} has been successfully applied to various video game environments to create human-level or even super-human agents~\cite{vinyals2019grandmaster, openai2019dota, ctf, vizdoom_competitions, dqn, ye2019mastering}, and show promise as a general way to teach computers to play games. However, these results are accomplished with a significant amount of engineering, including questions like:
    how should the agent perceive the environment,
    what should the rewards be and
    when should the game terminate, just to name a few examples. 
    One of these concerns is the \textit{action space}: 
    how does the agent act in the environment?
    Do we restrict the number of available actions? 
    Should we simplify it by creating combinations of actions? 
    % How about continuous controls like mouse movement?
    How do we deal with continuous controls like mouse movement?
    % Intuitively, learning to control a system with more buttons is harder (think how you would learn), but these modifications could limit agent's performance.
    Intuitively, learning to control a system with more buttons is harder, as the agent has to learn what each of the actions mean.
    Reducing the number of buttons might ease the learning but comes with the risk of limiting the agent's performance.
    %Such \textit{action space shaping}, modifications to the original action space with the goal of speeding up training and improving final performance we call \textit{action space shaping}.

    % Such \textit{action space shaping} is prominent in reinforcement learning competitions, especially in ones that use a video game as the environment. 
    Such \textit{transformations} of the action space, which we call \textit{action space shaping}, are prominent in reinforcement learning research and competitions, especially when it comes to video games.
    Environments like Doom~\cite{vizdoom} and Minecraft~\cite{johnson2016malmo} have large action spaces with dozens of buttons, and in related competitions \textit{all top participants} modified the actions~\cite{milani2020minerl, vizdoom_competitions}. 
    %This shaping comes in the forms of removing actions, combining different actions into one action and discretizing continuous actions, with the goal of making the learning easier for the agent, similar to \textit{reward shaping}~\cite{ng1999policy}.
    This action space shaping comes in the forms of removing actions, combining different actions into one action and discretizing continuous actions. The goal is to ease the learning for the agent, similar to \textit{reward shaping}~\cite{ng1999policy}.
    
    Along with the well-known work on mastering Starcraft II~\cite{vinyals2019grandmaster} and Dota 2~\cite{openai2019dota} with reinforcement learning, other games have received similar attention, such as modern first-person shooters~\cite{harmer2018imitation, ctf}, Minecraft~\cite{mineRLcompetition, johnson2016malmo}, popular fighting-games like Super Smash Bros~\cite{firoiu2017beating}, other massive online battle arenas (MOBAs)~\cite{ye2019mastering} and driving in GTA V~\cite{gao2018reinforcement}. All of these works do action space shaping, either because of limitations of the learning environment or because of the sheer number of actions, {\em e.g.} in strategy games and MOBAs. 
    %Even with seemingly simple-to-control games like Minecraft, actions can be represented as a \say{choose one from thousands}, or as a keyboard-like setup with \say{which buttons should be down}.
    
    The effect of different action spaces is no stranger to RL research.
    A large number of possible actions is known to lead to over-optimistic estimates of future rewards~\cite{thrun1993issues}. 
    Previous research has addressed this problem by removing un-promising actions~\cite{zahavy2018learn}, or by finding the closest neighbors of promising ones~\cite{dulac2015deep}. 
    Other works extended existing methods by adding support for different action spaces~\cite{branchingdqn, gu2016continuous} or by supporting large, complex action spaces~\cite{sc2}. 
    The work by Delalleau \textit{et al.}~\cite{delalleau2019discrete} shares our focus and mindset, where authors discussed different ways of processing complex action spaces, specifically for video games. However, this and other related works have not included experiments that study specific changes to the action-spaces. We fill this gap by running experiments on various video game environments, testing transformations successfully used in different competitions. Our core research question is \say{\textbf{do these transformations support the training of reinforcement learning agents?}}
    
    By answering this question in the context of video games, we aim to reduce the number of dials to tune when applying RL to games. 
    This is especially useful in the case of new environments, where it is unknown if RL agents can learn to play the game. 
    If an agent fails to learn, it is often unclear if the issue lies in the learning algorithm, the observation space, the rewards, the environment dynamics or the action space.
    By studying which types of action spaces work and which do not, we hope to remove one of these possibilities, making prototyping and further research easier.
    
    We start our contribution by summarizing different types of action space shaping done, recognizing common transformations and then verifying their effectiveness with empirical experiments. 
    We start with experiments in a toy-environment, and then move on to Atari, ViZDoom, Starcraft II and Obstacle Tower environments, to cover a broad range of different environments.

\section{Action space shaping}
    When applying RL to a specific task, it is common to use reward shaping~\cite{ng1999policy} to create a denser or more informative reward signal and thus making learning easier (agent's average score increases sooner in the training process). 
    We define action space shaping in a similar manner: modifying the original action space by using transformations, with the intent to make learning easier.
    By original action space, we refer to the action space provided by the learning environment authors or the one defined by the game. For many computer games, this would be the buttons on a keyboard and the movement of the mouse.
    
    \subsection{Reinforcement learning background}
        In reinforcement learning we consider agents that interact with environments at discrete time-steps $t$ by taking actions $a_t$ from a set of possible actions $\mathcal{A}$.
        Each step, an agent receives the current environment state $s_t$ from the set of all possible states $\mathcal{S}$ and a numerical reward $r_t$.
        The reward is given by a reward function $\mathcal{R}(s, a) = \mathbb{E}[r_{t+1} \mid s_t = s, a_t = a]$.
        The goal is then to find a policy $\pi(a \mid s) = \mathbb{P}[a_t = a \mid s_t = s]$ that maximizes the episodic reward in expectation $\mathbb E \left [ \sum_{t} r_t \right ]$, where expectation goes over the distribution of states and actions induced by the environment and the policy.
    
    \subsection{Types of action spaces}
        Similar to the action spaces established in the OpenAI Gym~\cite{gym}, we define the fundamental action spaces as follows:
        
        \begin{itemize}
            \item \texttt{Discrete}. Arguably the most used action space, where each action is an integer $a \in \{0, 1, \ldots N\}$, where $N \in \mathbb N$ represents the number of possibilities to choose an action from.
            For example, an agent playing the game Pong can choose between either \emph{move up}, \emph{move down} or \emph{stay}.  
            \item \texttt{MultiDiscrete}. An extension to \texttt{Discrete} space\footnote{\texttt{Discrete} is a special case of \text{MultiDiscrete}, but we separate the two as \texttt{Discrete} is used so wildly.}, where action $\mathbf a$ is a vector of individual discrete actions $a_i \in \{0, 1, \ldots N_i\}$, each with possibly different number of possibilities $N_i$.
            Arguably, this is closer to natural action spaces. 
            for example, a keyboard is a large \texttt{MultiDiscrete} space, where each \texttt{Discrete} variable can be either down or up.
            \item \texttt{Continuous}. Action $a \in \mathbb R$ is a real number/vector, rather than a discrete choice of many options. The amount of mouse movement~\cite{vizdoom, johnson2016malmo} or acceleration applied are \texttt{Continuous} actions, for example.
        \end{itemize}
        
        These action spaces can then be combined into more complex ones, where one action can consist of mixture of all of them, as described in~\cite{delalleau2019discrete}. A set of keyboard buttons and mouse control could be represented as a combination of \texttt{MultiDiscrete} and two \texttt{Continuous} actions, one continuous action per mouse movement axis, for example.
        
        \texttt{MultiDiscrete} spaces are often treated as independent \texttt{Discrete} decisions~\cite{delalleau2019discrete, kanervisto2019torille}. Policies for \texttt{Continuous} spaces have been implemented in various ways, that come with different advantages and disadvantages, one of which is the bounding of possible actions to a certain range~\cite{sac, chou2017improving}. Put quite simply, support for \texttt{Continuous} spaces is often harder to implement correctly than for \texttt{Discrete} spaces.
        
        %In practical implementations, \texttt{MultiDiscrete} spaces are often treated by treating each \texttt{Discrete} decision independent of each other~\cite{delalleau2019discrete, kanervisto2019torille}. \texttt{Continuous} actions can be implemented in various ways~\cite{sac, chou2017improving}, with various results. They are also susceptible to numerical issues~\cite{sac}. Quite simply, support for them is harder to implement correctly than for \texttt{Discrete} spaces.
        
        %\texttt{Discrete} is the most common out of all these action spaces, as it is easy to work with in theory and practice. In fact, in standard Markov Decision Process, actions are discrete choices. Q-learning and deep Q-learning~\cite{dqn} only work on \texttt{Discrete} action spaces by default, for example, but modifications for \texttt{MultiDiscrete} and \texttt{Continous} spaces exist~\cite{branchingdqn, gu2016continuous}. Common way to implement \texttt{MultiDiscrete} is to assume all the discrete variables are independent, and handle them separately~\cite{delalleau2019discrete, kanervisto2019torille}. Agents working on a \texttt{Continuous} space often parametrize a probability distribution, which requires selecting this probability distribution as well as limiting the variables to avoid singularity (\textit{e.g.} variance of a normal distribution should not be too low)~\cite{ppo, sac}.

    \subsection{Action space shaping in video game environments}
        \label{sec:transformations}
        
        \begin{table*}[] 
            \centering
            \begin{threeparttable}[b]
                \scriptsize
                \centering
                \begin{tabular}{llllll}
                \textbf{Environment Name} & \textbf{Original action space} & \textbf{Transformation} & \begin{tabular}[c]{@{}l@{}}\textbf{Transformed} \textbf{action space}\end{tabular} & \textbf{Performance} & \textbf{Reference} \\ \hline
                \multirow{5}{*}{MineRL} & \multirow{5}{*}{\begin{tabular}[c]{@{}l@{}}Multi-discrete(2, 2, 2, 2,\\2, 2, 2, 2, 7, 8, 5, 8, 3),\\ Continuous(2)\end{tabular}} & \colorbox{cyan}{DC} \colorbox{magenta}{RA} \colorbox{yellow}{CMD} & Discrete(36) &  & \tnote{1} \\
                 &  & \colorbox{cyan}{DC} \colorbox{magenta}{RA} \colorbox{yellow}{CMD} & Discrete(10) & \nth{1} place &~\cite{skrynnik2019hierarchical} \\
                 &  & \colorbox{cyan}{DC} \colorbox{magenta}{RA} \colorbox{yellow}{CMD} & Discrete(216) & \nth{2} place & ~\cite{milani2020minerl} \\
                 &  & \colorbox{cyan}{DC} \colorbox{magenta}{RA} & \begin{tabular}[c]{@{}l@{}}Multi-discrete(2, 2, 3,\\3, 7, 8, 5, 8, 3, 40)\end{tabular}  & \nth{3} place &~\cite{scheller2020sample} \\
                 &  & \colorbox{cyan}{DC} \colorbox{magenta}{RA} & \begin{tabular}[c]{@{}l@{}} Multi-discrete(2, 2, 2,\\5, 8, 3, 8, 7, 3, 3)\end{tabular} & \nth{5} place &~\cite{milani2020minerl} \\  \hline
                \multirow{2}{*}{\begin{tabular}[c]{@{}l@{}}Unity Obstacle Tower\\Challenge\end{tabular}} & \multirow{2}{*}{Multi-discrete(3, 3, 2, 3)} & \colorbox{magenta}{RA} \colorbox{yellow}{CMD} & Discrete(12) & \nth{1} place &~\cite{nichol_2019} \\
                 &  & \colorbox{magenta}{RA} & Discrete(6) & \nth{2} place & \tnote{2} \\  \hline
                \multirow{2}{*}{VizDoom (Doom)} & \multirow{2}{*}{\begin{tabular}[c]{@{}l@{}}38 binary buttons, \\ 5 continuous\end{tabular}} & \colorbox{magenta}{RA} \colorbox{yellow}{CMD} & Discrete(256) & \nth{1} place (Track 2) &~\cite{DK2017} \\
                 &  & \colorbox{magenta}{RA} & Discrete(6) & \nth{1} place (Track 1) &~\cite{Wu2017Doom} \\  \hline
                Atari & Discrete(18) & \colorbox{magenta}{RA} & Discrete(4 - 18) &  &~\cite{dqn} \\  \hline
                StarCraft II & \begin{tabular}[c]{@{}l@{}}Multi-discrete\end{tabular} & \colorbox{cyan}{DC} \colorbox{magenta}{RA} & Multi-discrete &  &~\cite{sc2, zambaldi2018deep, vinyals2019alphastar, vinyals2019grandmaster} \\  \hline
                Dota 2 & \begin{tabular}[c]{@{}l@{}}Multi-discrete\end{tabular} & \colorbox{cyan}{DC} \colorbox{magenta}{RA} & Multi-discrete &  &~\cite{openai2019dota} \\  \hline
                GTA V (car driving only) & Multi-discrete & \colorbox{magenta}{RA} \colorbox{yellow}{CMD} & Discrete(3) &  &~\cite{gao2018reinforcement} \\  \hline
                Torcs & Multi-discrete & \colorbox{magenta}{RA} \colorbox{yellow}{CMD} & Discrete(3) &  &~\cite{gao2018reinforcement, a3c} \\  \hline
                %MUJOCO humanoid & Continuous(23) & \colorbox{cyan}{DC} & \todo{Do we want to include this?}  &  &~\cite{tang2019discretizing} \\  \hline
                DMLab (Quake 3) & \begin{tabular}[c]{@{}l@{}}Multi-discrete(3, 3, 2, 2, 2),\\ Continuous(2)\end{tabular} & \colorbox{magenta}{RA} \colorbox{yellow}{CMD} & Discrete(9) &  &~\cite{dmlab,impala2018} \\  \hline
                Honor of Kings (MOBA) & \begin{tabular}[c]{@{}l@{}}Multi-Discrete, \\ Continuous\end{tabular} & \colorbox{magenta}{RA} \colorbox{yellow}{CMD} & \begin{tabular}[c]{@{}l@{}}Multi-discrete,\\ Continous\end{tabular} &  &~\cite{ye2019mastering} \\  \hline
                Little Fighter 2 (lf2gym) & Multi-Discrete(2,2,2,2,2,2,2) & \colorbox{yellow}{CMD} & Discrete(8) &  &~\cite{li2018deep} \\  \hline
                \end{tabular}
                \caption{Summary of action space shaping done in different video game-based competitions and learning environments. By ``original" space, we refer to action space originally provided by the environment designers. ``Multi-discrete($\cdot$)" shows the number of discrete variables, and number of choices for each. DC: Discretize continuous actions, RA: Remove actions, CMD: Convert multi-discrete to discrete.}
                \label{tab:actionSpaceTransformations}
                \begin{tablenotes}
                    \item [1] https://github.com/minerllabs/baselines/tree/master/general/chainerrl
                    \item [2] https://slideslive.com/38922867/invited-talk-reinforcement-learning-of-the-obstacle-tower-challenge
                \end{tablenotes}
            \end{threeparttable}
        \end{table*}
        
        Table \ref{tab:actionSpaceTransformations} summarizes action space shaping done by top-participants of different video game competitions and authors using video game environments for research. 
        In this section, we give an overview of the three major categories of action space transformations found throughout these works.
        
        \paragraph{\colorbox{magenta}{RA} Remove actions} 
            Many games include actions that are unnecessary or even harmful for the task. In Minecraft~\cite{mineRLcompetition, johnson2016malmo}, the \say{sneak} action is not crucial for progressing in the game, and therefore is often removed. The action \say{backward} is also often removed~\cite{nichol_2019, milani2020minerl}.
            Otherwise, the agent would waste training time by constantly switching between moving forward and backward, effectively jittering in place rather than exploring the environment. 
            % The removed action maybe be left on, which was done Minecraft MineRL competition baselines, where always executing \say{attack} helped the agents to learn gathering resources.
            Removed actions maybe set to \say{always on}, which was a popular transformation in the Minecraft MineRL competition, where always executing \say{attack} helped the agents to learn gathering resources~\cite{milani2020minerl, skrynnik2019hierarchical, scheller2020sample}.
            
            %An action can be removed by setting it to either always off or always on.
            %The latter was done in the Minecraft MineRL competition baselines, where always \say{attacking} helped the agents to learn gathering resources.
            
            Reducing the number of actions helps with exploration, as there are less actions to try, which in return improves the sample efficiency of the training.
            %as there are less actions to try.
            However, this requires domain knowledge of the environment, and it may restrict agent's capabilities.
            
            %In our experiments, we will use term \say{minimal} to refer to action spaces that contain only the necessary actions to complete the task.
        
            % Motivation: smaller action space eases the exploration problem (less actions to try out) and therefore speeds up training
            % Tradeoff: needs human priors on what acitons are necessary, danger of droping necessary actions
        
        \paragraph{\colorbox{cyan}{DC} Discretize continuous actions} 
            Many environments include \texttt{Continuous} actions, \textit{e.g.} in the form of mouse movement or camera turning speed.
            % Users of RL often discretize this either by splitting it into a set of bins, or by defining three discrete choices: Negative, zero or positive. 
            These actions are often discretized, either by splitting them into a set of bins, or by defining three discrete choices: negative, zero and positive.
            This is especially common with camera rotation, where agents can only choose to turn the camera left/right and up/down at a fixed rate per step~\cite{skrynnik2019hierarchical, milani2020minerl, impala2018, scheller2020sample}. 
            A downside is that this turning rate is a hyper-parameter, which requires tuning. 
            If the rate is too high, the actions are not fine-grained, and the agent may have difficulties in aiming at a specific spot. 
            If too small, it may slow down the learning or lead to sub-optimal behaviour as it takes more steps to aim at a specific target. 
            
            % Motivation: small number of discrete actions easier to learn than continuous. (why is this the case?)
            % Tradeoff: less finegrained actions, restricts agent to a action subspaces
        
        \paragraph{\colorbox{yellow}{CMD} Convert multi-discrete actions to discrete}
            Especially in ViZDoom~\cite{vizdoom_competitions} and Minecraft~\cite{mineRLcompetition}, it is common to turn \texttt{MultiDiscrete} actions into a single \texttt{Discrete} action, with all possible combinations of the \texttt{MultiDiscrete} actions. 
            Since the resulting action space combinatorially explodes quickly with an increasing \texttt{MultiDiscrete} space, this is usually combined with removing some of the actions. 
            This can be either dropping unnecessary actions as described above, manually selecting the allowed combinations (as done in MineRL~\cite{skrynnik2019hierarchical, milani2020minerl}) or by limiting maximum number of pressed buttons (as done in ViZDoom~\cite{DK2017, Wu2017Doom}). 
            
            This transformation is intuitively motivated by the assumption that it is easier to learn a single large policy than multiple small policies, as well as technical limitations of some of the algorithms. For example, methods like Q-learning~\cite{dqn} only work for \texttt{Discrete} action spaces. While a modified version of Q-learning exists for \texttt{MultiDiscrete} spaces~\cite{branchingdqn}, this is not commonly used.

\section{Experiments}
    
    %\todo{Consistency: Italics for environment names, emphasis action-spaces and texttt action-space types}

    With the major action-space transformations summarized above, we move onto testing if these transformations are truly helpful for the learning process. We do this by training RL agents in a larger variety of environments and comparing the learning process between different action-spaces.

    Our main tool for evaluation are learning curves, which show how well agents perform at different stages of training. These show the speed of learning (how fast curve rises), show the final performance (how high curve gets) and if the agent learns at all (if the curve rises). We will use four different games (Doom, Atari, Starcraft II and Obstacle Tower), along with a toy-environment. Source code for the experiments is available at \url{https://github.com/Miffyli/rl-action-space-shaping}.

    \subsection{Reinforcement learning agent}
        We use the \textit{Proximal Policy Optimization} (PPO)~\cite{ppo} algorithm for training the agents.
        We employ the high-quality implementations from stable-baselines~\cite{stable-baselines} and rllib~\cite{liang2018rllib}, which support various action-spaces. We opt for PPO rather than other recent state-of-the-art algorithms~\cite{sac, a3c} for its simplicity, previous results in the environments used in this work and maturity of their implementations. We do not expect final insights to change between different learning algorithms, as action-space transformations are not part of the algorithm design, but part of the environment. An exception to this are the \texttt{Continous} actions, which have multiple ways to implement them, and come with additional parameters to tune~\cite{sac, continuous}.
        
        Unless otherwise noted, we will use the following hyper-parameters: eight parallel environments, from each of which we collect $256$ before running four epochs of updates on the gathered data. Entropy coefficient/weight is set to $0.01$, and PPO clipping to $0.2$. Network is optimized with Adam~\cite{adam} with learning rate $2.5 \cdot 10^{-4}$.

    \subsection{Get-To-Goal experiments}
        For rapid experimentation with different action spaces and their effects on the learning performance, we implement a simple reach-the-goal environment. 
        The environment consists of a bounded 2D area, a player and a goal. 
        The game starts with a player and a goal at random locations and ends when the player either reaches the goal (reward $1$) or when environment times out (reward $0$). 
        Agent receives a 2D vector pointing towards the goal, as well as their current heading as a $(\cos(\phi), \sin(\phi))$ tuple, where $\phi \in [0, 2 \pi]$ is the relative rotation angle.
        We use this environment to test \colorbox{cyan}{DC} by using discrete and continuous variants of the action space.
        
        \begin{itemize}
            \item \textbf{Multi-Discrete}: Player can move on two axes with the four buttons \textit{Up}, \textit{Down}, \textit{Left} and \textit{Right} (\texttt{MultiDiscrete}).
            \item \textbf{Discrete}: A flattened version of above, where only \textit{one} of the buttons may be pressed at a time, \textit{i.e.} no diagonal movement allowed (\texttt{Discrete}).
            \item \textbf{Continuous}: Player specifies the exact direction of the next move with a continuous value, with $0$ representing straight up, $90$ straight right and $180$ straight down (\texttt{Continuous}).
            \item \textbf{Tank, Discrete/Multi-Discrete}: Player has a heading $\phi$, and it can choose to increase/decrease it (turn left/right), and/or to move forward/backward towards the heading (\texttt{Discrete} and \texttt{MultiDiscrete} versions).
        \end{itemize}
        
        To study \colorbox{magenta}{RA} and \colorbox{yellow}{CMD}, we train agents with different number of actions available in \texttt{Discrete} and \texttt{MultiDiscrete} spaces.
        Each action moves the player to a different direction, equally spaced on a unit circle (\textbf{Extra actions}). 
        We also run experiments with additional no-op actions, which do not do anything, to simulate \textit{e.g.} the \texttt{USE} action in Doom, which only works in specific scenarios (\textbf{Bogus actions}). We also test the effect of \say{backward} and \say{strafe} actions, which are often removed in FPS games, by enabling/disabling them in tank-like action spaces. All experiments are run with stable-baselines \cite{stable-baselines}.
        
        \begin{figure}[t]
            \centering
            \includegraphics[width=0.98\columnwidth]{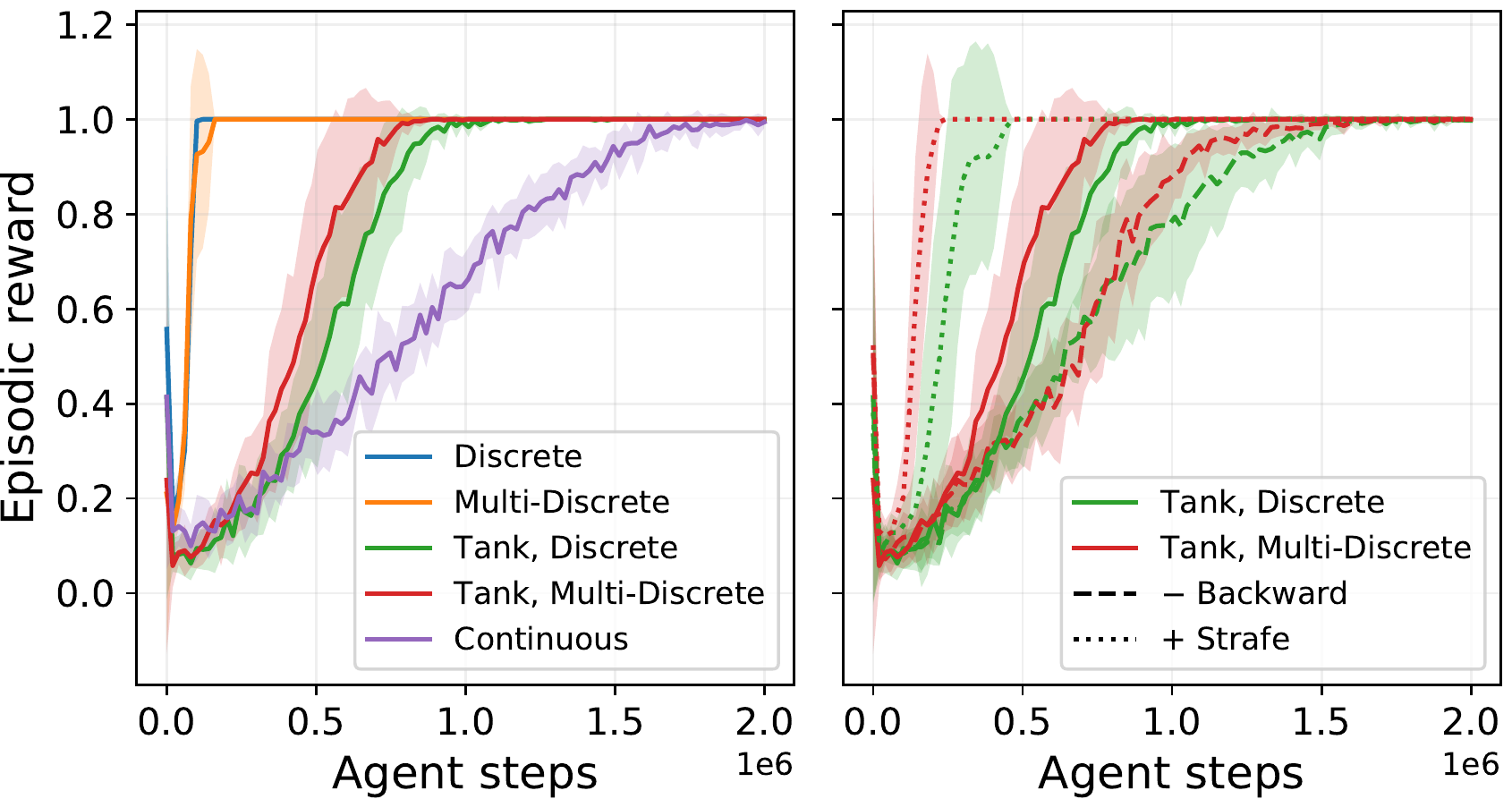}
            \caption
            {
                Learning curves in Get-To-Goal environment, with various action spaces (left), and various buttons available in tank-like controls (right). Averaged over ten repetitions, with shaded region representing the standard deviation.
            }
            \label{fig:gettogoal-results}
        \end{figure}
        
        \begin{figure}[t]
            \centering
            \includegraphics[width=0.98\columnwidth]{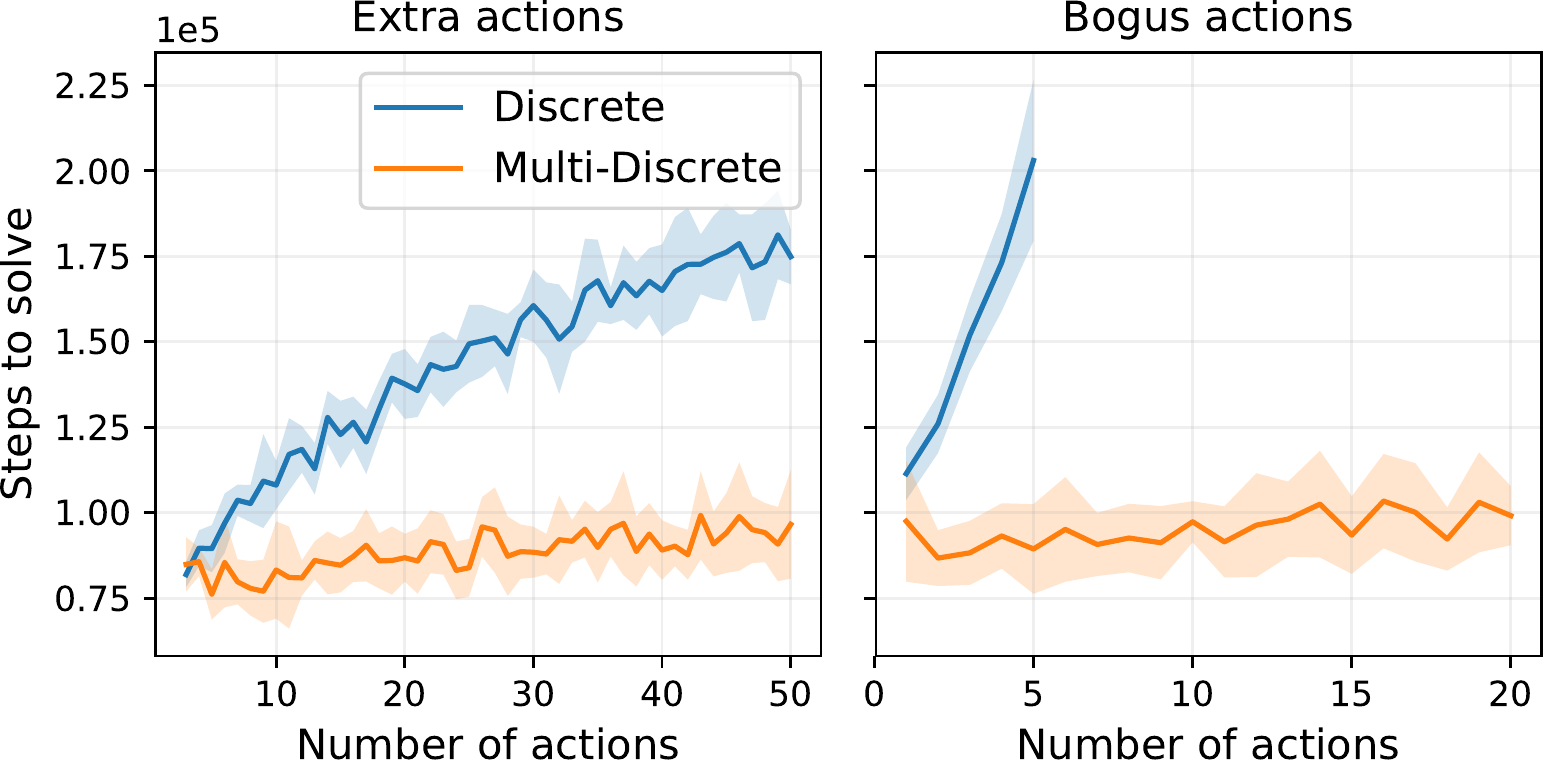}
            \caption{
                Results of an increasing number of actions in Get-To-Goal environment. With \textbf{Extra actions}, each action moves the player to a different direction, while in \textbf{Bogus actions} the extra actions do nothing. Averaged over ten repetitions.
            }
            \label{fig:gettogoal-extra-actions}
        \end{figure}
        
        Figure \ref{fig:gettogoal-results} (left) shows the results with different action-spaces. Learning with tank-like controls is slower than with the non-tank controls and learning with \texttt{Continuous} spaces is the slowest. It should be noted that with rllib \cite{liang2018rllib} we observed similar results, except \texttt{Continuous} learned faster than tank-like controls. This indicates that policies for \texttt{Continuous} actions are sensitive to the implementation, with the discrete options being robust to this. 
        
        Figure \ref{fig:gettogoal-results} (right) shows results with tank-like controls, with and without backward and strafe actions. 
        In both action-spaces, agents learn slower the more actions they have available. This demonstrates how removing actions can be detrimental to the performance (\colorbox{magenta}{RA}).
        It is evident, that agents learn faster on \texttt{MultiDiscrete} spaces than on the \texttt{Discrete} alternatives. % Combined with results in Figure \ref{fig:gettogoal-extra-actions}, which show how \texttt{MultiDiscrete} spaces are more robust to additional actions, demonstrates how \texttt{MultiDiscrete} can be more beneficial than \texttt{Discrete} space (\colorbox{yellow}{CMD}). 
        Figure \ref{fig:gettogoal-extra-actions} shows how \texttt{MultiDiscrete} spaces are more robust to additional actions.
        These results demonstrate that there are situtaions where RL agents can profit from \texttt{MultiDiscrete} compared to \texttt{Discrete} spaces (\colorbox{yellow}{CMD}). 
    
    \subsection{Atari experiments}
        With Atari games, we test \colorbox{magenta}{RA} and \colorbox{yellow}{CMD} transformations. Atari games have been a standard benchmark for DRL algorithms~\cite{dqn, ppo, a3c}, where their action space is defined as a \texttt{Discrete} action space. By default, the action space consists of only the necessary actions to play the game (\textbf{minimal}). 
        % We compare these minimal actions against the full action space (which includes the \say{no-op} actions), and \texttt{MultiDiscrete} space, where the joystick an fire-button are additional buttons ($9$ options) and fire-button is another ($2$ options).
        We compare these minimal actions against the \textbf{full} action space, and a \textbf{multi-discrete} action space, where joystick and fire-button are additional buttons with $9$ and $2$ options respectively. 
        %The actions not included in the \textbf{minimal} space may still have an effect, e.g. in \textit{Enduro} game \say{UPLEFT} action moves car left, despite not included in the \textbf{minimal} space. Note that the \textbf{multi-discrete} space also includes all possible actions.
        
        We use six games, selected for varying number of actions in the minimal action space. All games have $18$ actions in the full space, while \textit{Space Invaders} and \textit{Q*bert} have six, \textit{MsPacman} and \textit{Enduro} have nine and \textit{Breakout} has four actions for minimal space. \textit{Gravitar} uses all $18$ actions, and thus do not have minimal spaces. We use the \say{v4} versions of the environments (e.g. \texttt{GravitarNoFrameskip-v4}), which are easier to learn. We use the PPO hyper-parameters from stable-baselines rl-zoo~\cite{rl-zoo}.

        \begin{figure}[t]
            \centering
            \includegraphics[width=0.98\columnwidth]{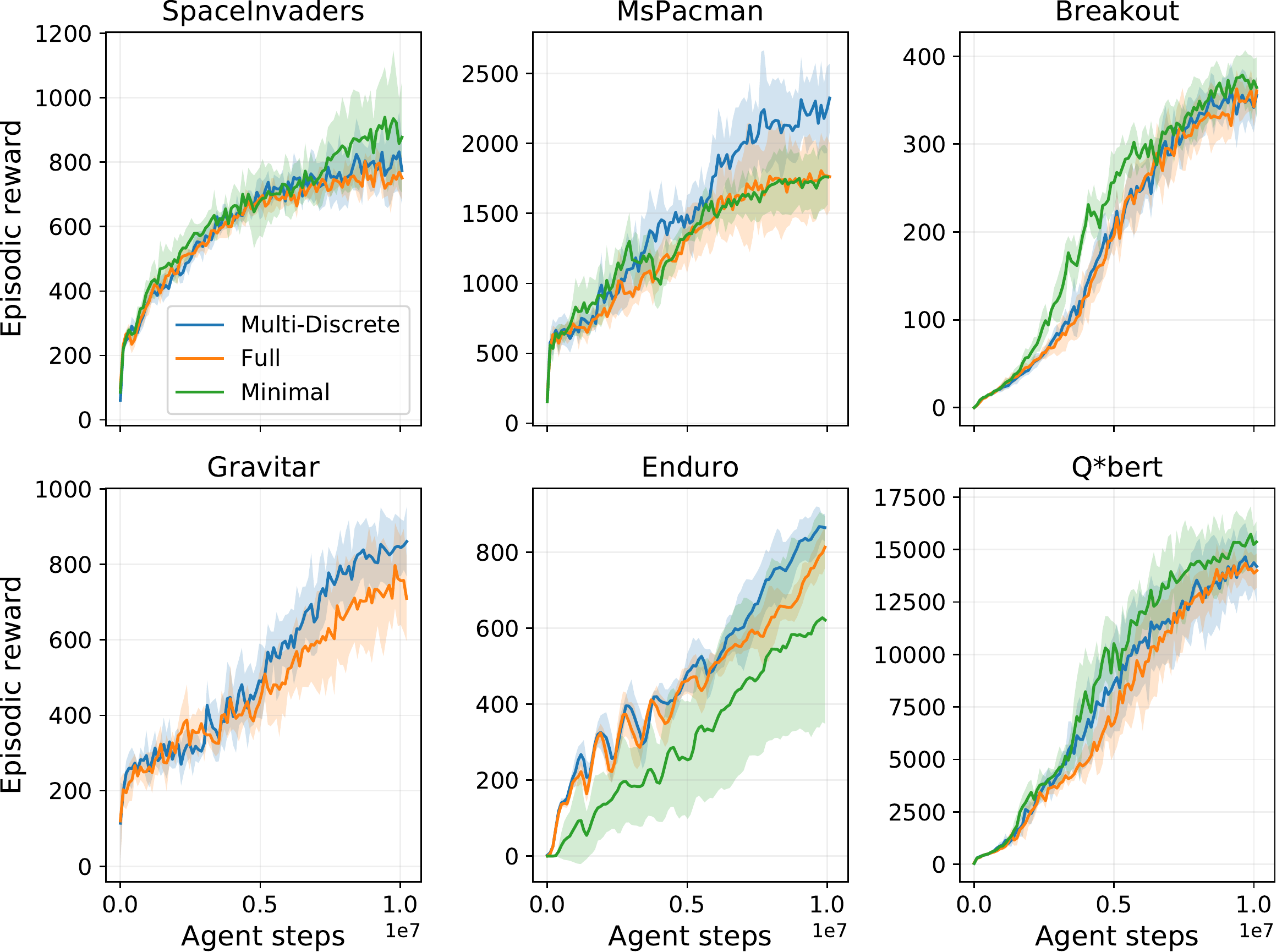}
            \caption{
                Results with six Atari games, using \textbf{minimal}, \textbf{full} and \textbf{multi-discrete} actions. Averaged over five training seeds. \textbf{Minimal} action-set only includes actions that are available in that specific game. \textbf{Multi-discrete} uses all possible actions, but by separating joystick and fire-button into two different discrete variables. \textit{Gravitar} does not have a \textbf{minimal} action space.
            }
            \label{fig:atari}
        \end{figure}
        
        Figure \ref{fig:atari} shows the resulting learning curves. On average, there is no clear difference between the different action-spaces over all games, with \textit{MsPacman} being an exception. Here the \textbf{multi-discrete} agent achieved almost one-quarter higher score than other action spaces. Interestingly, in \textit{Enduro}, both action spaces using all buttons out-perform the \textbf{minimal} space, despite the fact that full space does not offer any new actions for agent to use. With these results, removing actions (\colorbox{magenta}{RM}) can limit the performance, but overall does not change results. Same applies to converting multi-discrete into discrete (\colorbox{yellow}{CMD}), although in one of the games it did obtain higher performance.

    \subsection{Doom experiments}
        
        With Doom (1993), we test all of the transformations (\colorbox{cyan}{DC}, \colorbox{magenta}{RA} and \colorbox{yellow}{CMD}). We use the environment provided by ViZDoom interface~\cite{vizdoom},
        %which has been used in three separate competitions on competing reinforcement learning agents based on visual information~\cite{vizdoom_competitions}.
        %While a crude game by modern standards, it includes most of the mechanics of a modern first-person shooter game, and the challenges thereof.
        with three different scenarios: \textit{Get-To-Goal}, \textit{Health gathering supreme} (HGS) and \textit{Deathmatch}. 
        The first is an implementation of the toy task described earlier, in the form of a first-person shooter scenario, where the agent is tasked to reach a goal object in a room.
        The player receives a reward of $+1$ if it reaches the goal, $0$ otherwise, including at the timeout-termination after $2100$ frames (one minute of game-time). 
        \textit{HGS} and \textit{Deathmatch} are Doom scenarios, where the player has to gather medkits in a maze to survive (HGS) and fight against randomly spawning enemies (Deathmatch). We modify the Deathmatch scenario to give a $+1$ reward per kill. We also make enemies weaker, so they die from one shot. Otherwise, the scenario is too difficult to learn by a PPO agent in the given time. We test four sets of buttons (\colorbox{magenta}{RA}):
    
        \begin{itemize}
            \item \textbf{Bare-Minimum}. The only allowed buttons are moving forward, turning left and attack (Deathmatch). These are the bare-minimum number of buttons to complete tasks.
            \item \textbf{Minimal}. Same as \textbf{Bare-Minimum}, but with an option to turn right. This was a common configuration in \textit{e.g.} the MineRL competition~\cite{milani2020minerl}.
            \item \textbf{Backward}. Same as \textbf{Minimal}, but with the additional option of moving backward.
            \item \textbf{Strafe}. Same as \textbf{Backward}, but with the additional options of moving left and right. This corresponds to the original movement controls of Doom.
        \end{itemize}
        
        For each set of buttons, we test five different action spaces:
        The original \texttt{MultiDiscrete} space, where pressing each button down is its own discrete choice, three levels of discretization (\colorbox{yellow}{CMD}) and continuous mouse control (\colorbox{cyan}{DC}). Discretization is done by creating all possible combinations of pressed down buttons, with varying levels of buttons pressed down ($n=1$, $n=2$ or all). %This is common in ViZDoom competitions~\cite{vizdoom, vizdoom_competitions}.
        For continuous actions, \textbf{mouse} actions correspond to a \texttt{MultiDiscrete} action space, where turning left and right has been replaced with a mouse control (a scalar, representing how much we should turn left or right). This action space is not combined with the bare-minimum button-set. 
        
        Observations consist of grayscale (\textit{GetToGoal} and \textit{HGS}) or RGB (\textit{Deathmatch}) image of size $80 \times 60$, along with any game-variables enabled in the scenario. 
        %No frame-stacking is used, and agent only receives a single frame in all scenarios. 
        Each action is repeated for four frames. All experiments except \textbf{mouse} are run using stable-baselines, as only rllib supports mixed action-spaces. Other results are same between rllib and stable-baselines.
        
        \begin{figure*}[t]
            \centering
            \includegraphics[width=0.98\linewidth]{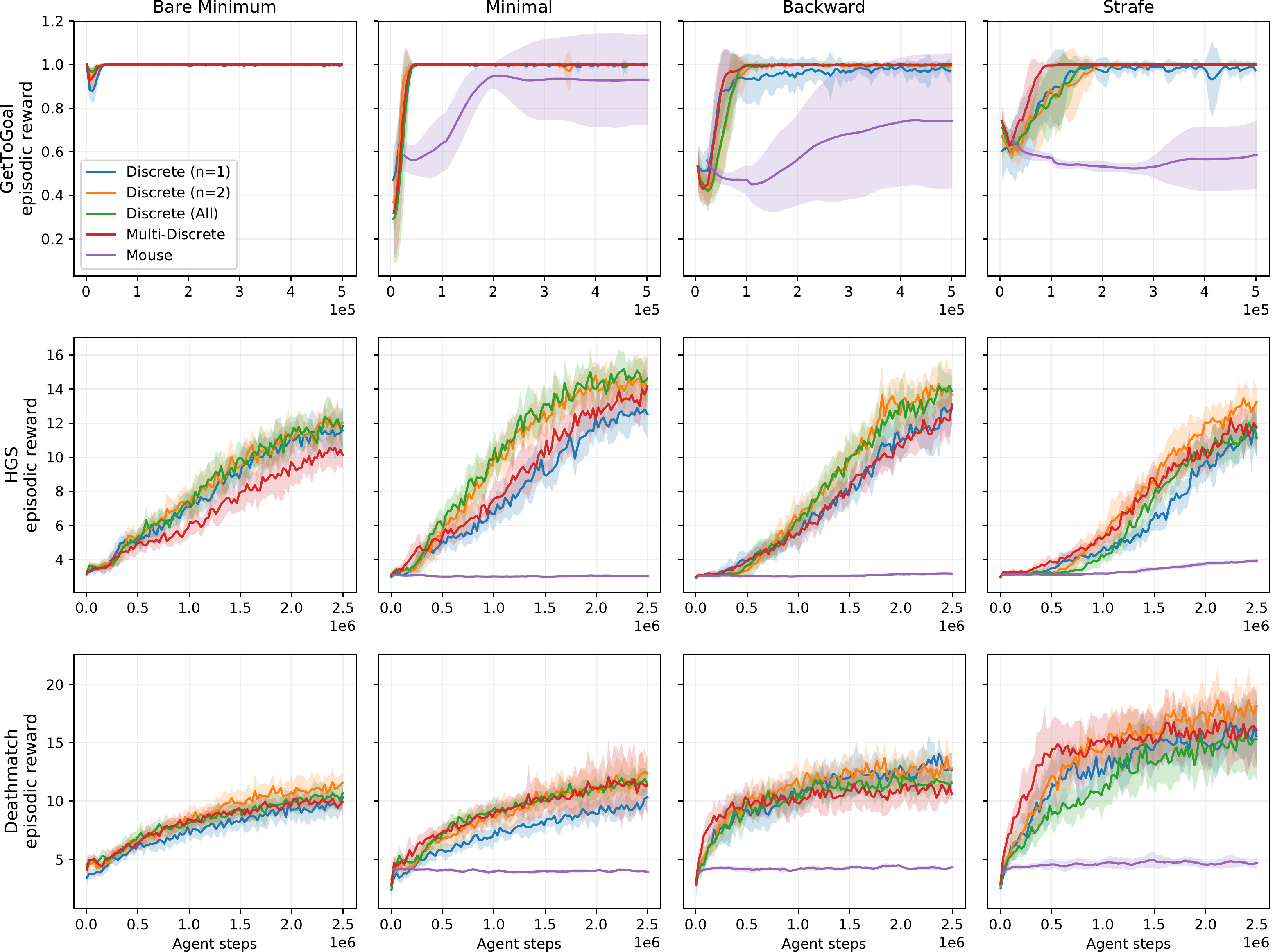}
            \caption{
                Results with the ViZDoom environment, with three environments (rows) and four sets of buttons (columns). Number of available actions increases from left to right. Curves are averaged over ten repetitions.
            }
            \label{fig:vizdoom-results}
        \end{figure*}
        
        Figure \ref{fig:vizdoom-results} shows the results. \textbf{Multi-discrete} action space performs as well as discretized versions. Using discrete actions with only one button down is the least reliable out of discrete spaces, as it is not able to reliably solve the easiest environment (\colorbox{yellow}{CMD}). Using continuous actions prevents learning in all but the simplest environment (\colorbox{cyan}{DC}). Increasing number of available actions improves the results in more difficult scenarios (\colorbox{magenta}{RA}).

    \subsection{Obstacle Tower experiments}

        Obstacle Tower~\cite{juliani2019obstacle} is a 3D platformer game with randomly generated levels, designed for reinforcement learning research. Its original action space is defined as a \texttt{MultiDiscrete} space, with options to move forward/backward and left/right, turn left/right and jump. We use this environment to test \colorbox{yellow}{CMD} and \colorbox{magenta}{RA} transformations, by disabling strafing, moving backward or forcing moving forward. Similar to Doom experiments, \texttt{Discrete} space is obtained by creating all possible combinations of the \texttt{MultiDiscrete} actions. Observations are processed using the \say{retro} setting of the environment ($84 \times 84$ RGB images). We use $32$ concurrent environments, set entropy coefficient to $0.001$ and collect $128$ steps per environment per update. These parameters were selected based on the previous experiments with Obstacle Tower environment.
        
        \begin{figure}[t]
            \centering
            \includegraphics[width=0.98\columnwidth]{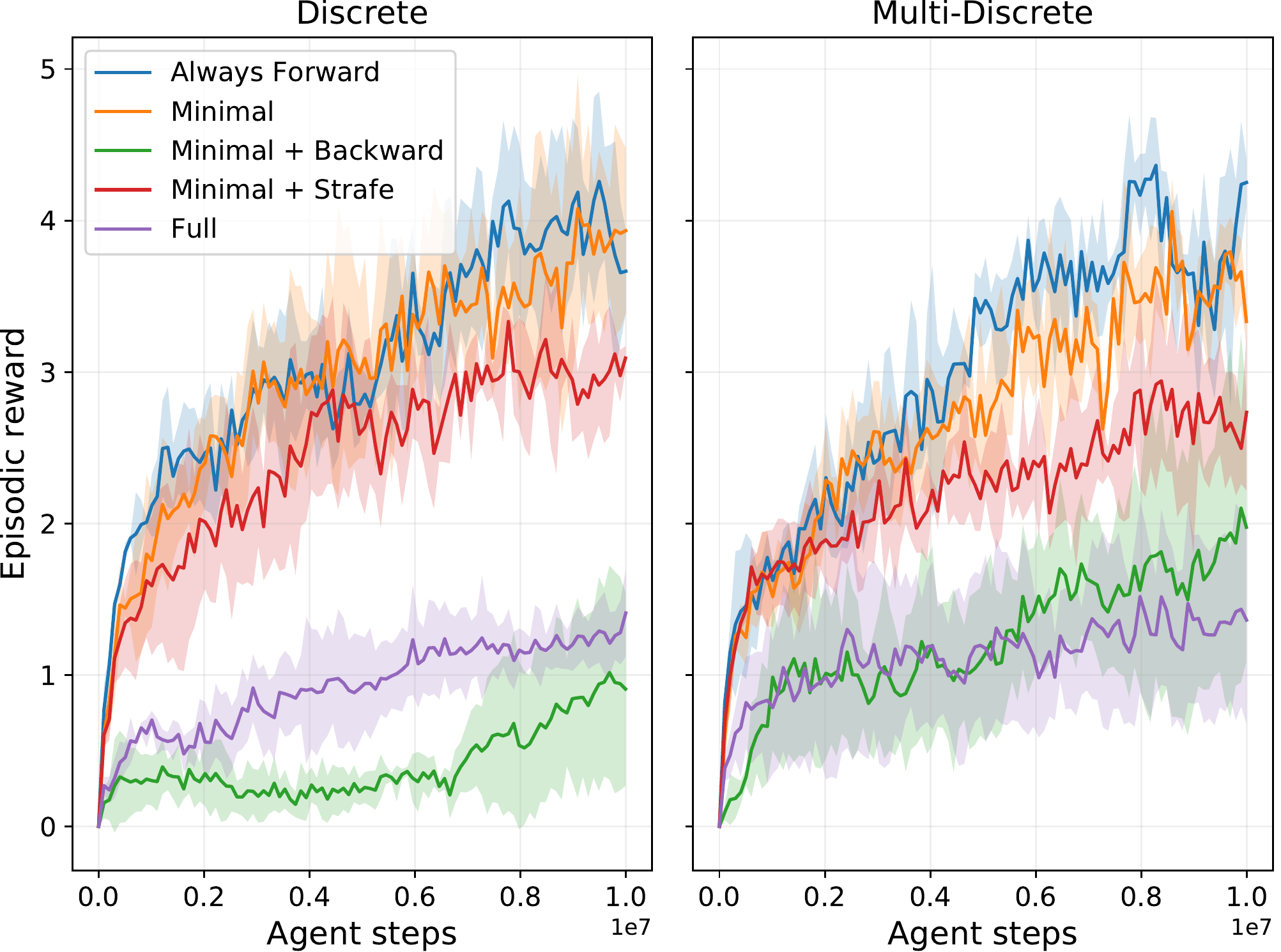}
            \caption{
                Results in the Obstacle Tower environment, various button sets and both \texttt{Discrete} and \texttt{MultiDiscrete} action spaces. \textbf{Always Forward} includes actions for turning and jumping. \textbf{Minimal} lets agent to choose when to move forward. Curves are averaged over three seeds.
            }
            \label{fig:ot}
        \end{figure}
        
        Figure \ref{fig:ot} shows the results. There is no significant difference between \texttt{Discrete} and \texttt{MultiDiscrete} spaces (\colorbox{yellow}{CMD}). The only major difference is with \textbf{backward} action: all experiments that allow moving backward have slower learning than the rest. This supports the intuition to remove unnecessary actions, and especially when the action can negate other actions during random exploration (\colorbox{magenta}{RA}).

    \subsection{StarCraft II experiments}
    
        % StarCraft II (SC2) is a complex multi-player real-time strategy game and a challenging domain for RL algorithms.
        % The game is played on a large variety of 2D maps, which due to the fog of war are only partly observable. 
        % Players must control hundreds of units and buildings, come up with long term strategies and make quick and precise decisions in order to defeat their opponent.
        % For humans the game is seen as one of the most challenging real-time strategy games.
        
        StarCraft II is a complex multi-player real-time strategy game and a challenging domain for RL algorithms.
        Particularly interesting for our work is the vast size of the action space.
        For environment interaction we use the StarCraft II learning environment (SC2LE)~\cite{sc2} that provides a parametric action space, with base actions and action parameters.
        Base actions describe an intended action such as \emph{move screen} or \emph{build barracks}.
        Action parameters such as \emph{screen coordinates} or \emph{unit id} modify these base actions.
        There are hundreds of base actions and depending on the base action and screen resolution up to millions or even billions of possible parameter combinations.
        
        % To investigate action space shapings for real-time strategy games like StarCraft II, we look at four out of the seven mini-games provided by SC2LE.
        We use four of the seven provided mini-games in our experiments: 
        %In this work, we investigate action space shaping for real-time strategy games like StarCraft II on the example of four out of the seven mini-games provided by SC2LE:
        % Each of these mini-games is targeted at a certain aspect of the full StarCraft II game:
        % We namely consider \emph{CollectMineralShards}, \emph{DefeatRoaches}, \emph{CollectMineralsAndGas} and \emph{BuildMarines}:
        \begin{itemize}
            \item \emph{CollectMinerlShards (CMS)} is a simple navigation task, where the player controls two Marines and must collect randomly scattered mineral shards as quickly as possible.
            \item In \emph{DefeatRoaches (DR)}, the player controls a small army of Marines and must defeat a small Roach army.
            \item The goal of \emph{CollecMineralsAndGas (CMAG)} is to build up a working economy and collect as much minerals and vespene gas as possible.
            \item \emph{BuildMarines (BM)} is targeted at establishing a unit production with the goal of building as many Marines as possible.
        \end{itemize}
        
        % \emph{CMS} and \emph{BR} focus on the precise control of multiple units (micro management). \emph{CMAG} and \emph{BM} are hard exploration tasks that focus on economical aspects of StarCraft II (macro management) and require credit assignment over long time horizons.
        
        To study \colorbox{magenta}{RA} on these mini-games, we test the effects of \textbf{Masked} and \textbf{Minimal} transformations. We also evaluate auto regressive policies (\textbf{AR}), which do not transform the action space but are of interest when dealing with large parametric action spaces:
        
        \begin{itemize}
            \item \textbf{Masked.} 
            In StarCraft II, at any given time, only a subset of all actions is available.
            To prevent agents from selecting unavailable actions and to ease the learning, base action policies are often masked by setting the probabilities of unavailable actions to zero~\cite{sc2,zambaldi2018deep}.
            % This prevents agents from choosing unavailable actions and should ease the learning.
            % We consider the masking of unavailable actions as a form of action space shaping, where we drop unnecessary actions.
            We run experiments with agents that are either equipped with action masking or not.
            Selecting an unavailable action results in a no-op action. 
            \item \textbf{Minimal.} 
            Not all actions are required to play the selected mini-games optimally.
            We evaluate the impact of removing unnecessary actions from the action space. 
            For each mini-game, we define a \emph{minimal} set of base actions required for optimal control.
            \item \textbf{AR.} 
            % Given the parameterized nature of the StarCraft II \texttt{MultiDiscrete} action space, optimal action parameters depend on the choice of the base action and other action parameters.
            The choice of optimal action parameters depends on the choices of base action and other action parameters.
            This is usually addressed by introducing an order of dependency and condition action parameter policies accordingly in an \emph{auto-regressive} manner~\cite{sc2, zambaldi2018deep, vinyals2019grandmaster}.
            % A common approach to handle dependent action parameters is to introduce an order of dependency and condition action parameter policies accordingly in an \emph{auto-regressive} manner~\cite{sc2, zambaldi2018deep, vinyals2019grandmaster}.
            % If converted into a flat \texttt{Discrete} action space with all possible combinations, the size will become untraceably large.
            % A common approach to handle dependent action parameters for such large action spaces is to introduce an order of dependency and condition action parameter policies accordingly in an \emph{auto-regressive} manner~\cite{sc2, zambaldi2018deep, vinyals2019grandmaster}.
            To study the effect of \emph{auto-regressive} policies, we run experiments where we condition action parameter policies on the sampled base action.
            Following~\cite{zambaldi2018deep}, we embed sampled base actions into a 16-dimensional continuous space and feed them as additional input to each of the action parameter policy heads.
        \end{itemize}
        
        % In this model, base actions and action parameters are represented by individual policy heads on top of a shared neural network.
        % For spatial action parameter policies like screen coordinates, there is a fully convolutional path through the network.
        For training, we employed IMPALA~\cite{impala2018}, a distributed off-policy actor critic method.
        % In IMPALA, the acting and learning is decoupled which allows large distributed setups and effective usage of GPUs.
        % To account for the resulting lag between action sampling and training step, the method introduces the \emph{V-trace} off-policy correction algorithm.
        % For the StarCraft II experiments, 
        We chose IMPALA over PPO for its scalability to large number of actors and since its strong performance on the SC2LE mini-games~\cite{zambaldi2018deep}. We use ControlAgent architecture and hyper-parameters described in~\cite{zambaldi2018deep}, except with fixed learning rate $10^{-4}$ and entropy coefficient $10^{-3}$.
        % All experiments were conducted with a batch size of 32, an unroll length of 80, a learning rate of \num{1e-4} and an entropy cost of \num{1e-3}. 
        %We represent policies with the ControlAgent architecture described in~\cite{zambaldi2018deep}, which is specifically targeted at the StarCraft II environment.
        %We use the same hyper-parameters as in~\cite{zambaldi2018deep}, with fixed learning rate of \num{1e-4} and an entropy coefficient of \num{1e-3}
        
        \begin{figure}[t]
            \centering
            \includegraphics[width=0.98\columnwidth]{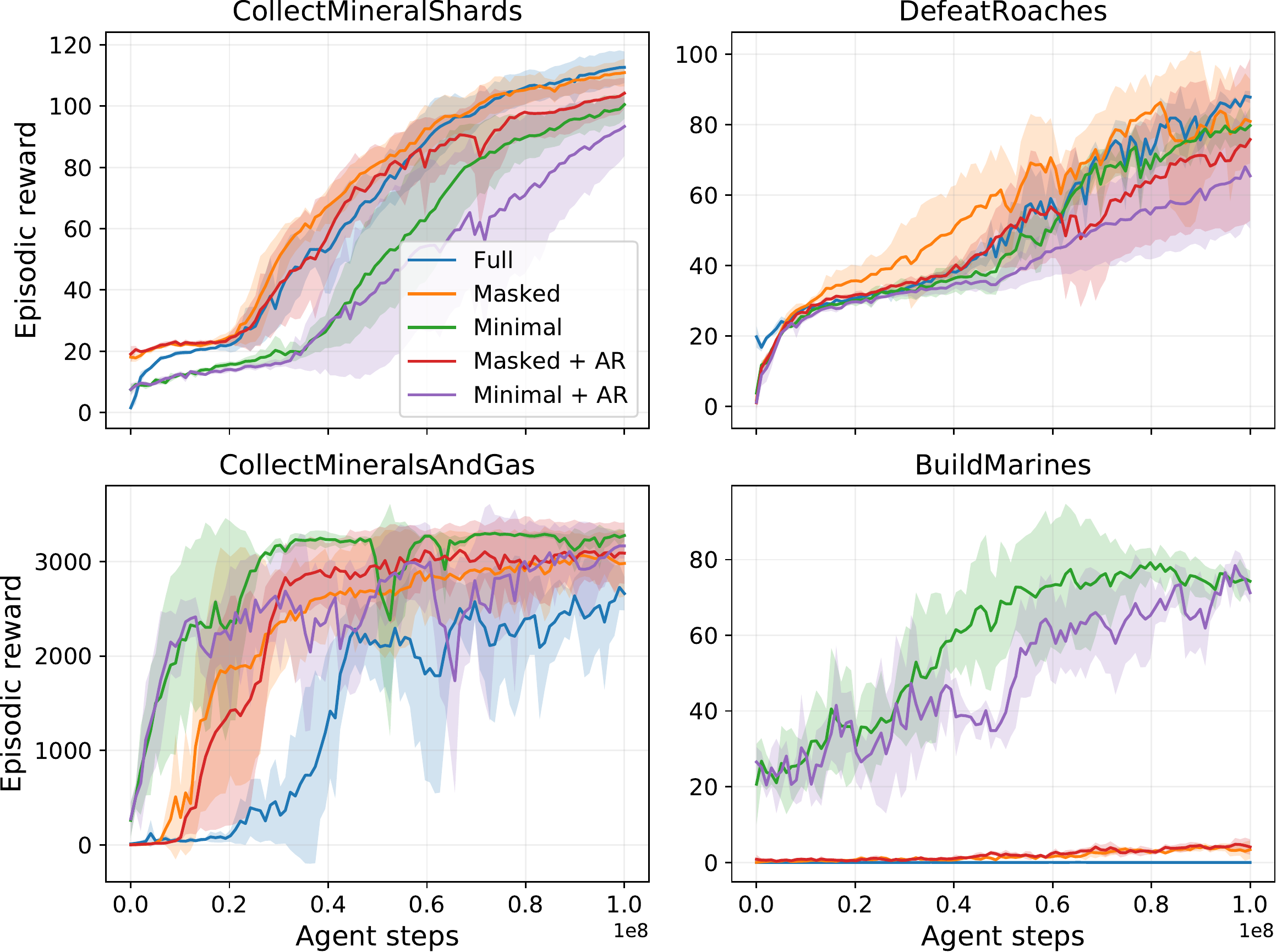}
            \caption{
                Results on StarCraft II mini games. \textbf{Full} includes all possible actions, \textbf{Masked} uses action-masking to remove unavailable actions and \textbf{Minimal} extends on this by removing all but necessary actions. We include experiments with autoregressive policies (\textbf{AR}), as they are commonly used with Starcraft II environment.
            }
            \label{fig:sc2-training-curves}
        \end{figure}
        
        The results are shown in Figure~\ref{fig:sc2-training-curves}. 
        Masking unavailable actions (\textbf{Masked}) turned out to be crucial for learning on \emph{BM} and significantly improved performance on \emph{CMAG}.
        For \emph{CMS} and \emph{DR} we did not see any improvement with masked polices.
        It is evident that agents trained on \emph{CMS} and \emph{DR} did not profit from minimal action spaces (\textbf{Minimal}).
        In contrast, on \emph{CMAG} and \emph{BM}, minimal action spaces improved the performance even for random policies at the beginning of training.
        Agents trained on \emph{CMAG} showed much quicker training progress and required less samples to achieve the final performance.
        On \emph{BM}, we see how \colorbox{magenta}{RA} can lead to significant improvements in the final performance of RL agents.
        With auto-regressive policies (\textbf{AR}), we did not observe significant improvement.
        We see that, within the limited number of samples, the agents learned only simple, sub-optimal behaviour, where they choose between few actions with distinct parameters.
        We believe that auto-regressive policies can be beneficial for learning better policies with larger set of actions.

\section{Discussion and conclusions}
    Overall, our results support the use of the action space transformations listed in Section \ref{sec:transformations}, with the exception being converting \texttt{MultiDiscrete} to \texttt{Discrete} spaces (\colorbox{yellow}{CMD}). 
    Removing actions (\colorbox{magenta}{RA}) can lower the overall performance (Doom), but it can be an important step to make environments learnable by RL agents (SC2, Obstacle Tower). 
    Continuous actions are harder to learn than discrete actions (Get-To-Goal) and can also prevent learning all-together (ViZDoom).
    Discretizing them (\colorbox{cyan}{DC}) improves performance notably.
    
    As for the \texttt{MultiDiscrete} spaces, we did not find notable difference between results with the \texttt{MultiDiscrete} and \texttt{Discrete} variants. Experiments on the Get-To-Goal task show how \texttt{MultiDiscrete} spaces scale well with an increasing number of actions, while \texttt{Discrete} do not.
    In all other environments (ViZDoom, Obstacle Tower, Atari) we observe no significant difference between the two.
    
    In this work, we have formalised the concept of \emph{action space shaping} and summarized its application in the previous RL research. We found three major transformations used throughout such work: removing actions, discretizing continuous actions and discretizing multi-discrete actions. We evaluated these transformations and studied their implications on five environments, which range from simple navigation tasks up to complex 3D first-person shooters and real-time strategy games.
    
    Answering the question presented in introduction, \say{do these transformations help RL training}, removing actions and discretizing continuous actions can be crucial for the learning process. Converting multi-discrete to discrete action spaces has no clear positive effect and can suffer from poor scaling in cases with large action spaces. Our guide for shaping an action space for a new environment is thus as follows:
    
    \textit{Start by removing all but the necessary actions and discretizing all continuous actions. Avoid turning multi-discrete actions into a single discrete action and limit the number of choices per discrete action. If the agent is able to learn, start adding removed actions for improved performance, if necessary.}
    
    In the future, we would like to extend this work with a more pin-pointed approach on what exactly makes the learning easier, both in theory (\textit{e.g.} exploration vs. number of actions) and in practice (\textit{e.g.} what kind of actions in games are bad for reinforcement learning). A simpler extension would be to repeat these experiments with more complex games like Minecraft, that have a large variety of mechanics and actions. Specifically, continuous actions serve more attention, along with combinations of different action-spaces. Finally, this work is but a steppingstone in the path towards automated action space shaping. We now know we can ease the learning process significantly with heuristics and manual engineering, and next we would like to see this process automated, \textit{e.g.} as part of the reinforcement learning process.

%\section*{Acknowledgements}
%   \todo{Something here, or is this stuff in the beginning?}

\bibliographystyle{ieeetr}
\bibliography{main}

\vspace{12pt}

\end{document}